\documentclass{article} 
\usepackage{iclr2026_conference,times}

\usepackage{amsmath,amsfonts,bm}

\def\eqref#1{equation~\ref{#1}}

\def\1{\bm{1}}

\def\va{{\bm{a}}}
\def\vb{{\bm{b}}}

\def\vg{{\bm{g}}}

\def\vk{{\bm{k}}}

\def\vo{{\bm{o}}}

\def\vq{{\bm{q}}}

\def\vv{{\bm{v}}}

\def\vx{{\bm{x}}}
\def\vy{{\bm{y}}}

\def\mK{{\bm{K}}}

\def\mQ{{\bm{Q}}}

\def\mV{{\bm{V}}}
\def\mW{{\bm{W}}}

\DeclareMathAlphabet{\mathsfit}{\encodingdefault}{\sfdefault}{m}{sl}
\SetMathAlphabet{\mathsfit}{bold}{\encodingdefault}{\sfdefault}{bx}{n}

\usepackage{hyperref}
\usepackage{url}

\usepackage{hyperref}
\usepackage{url}
\usepackage{graphicx}
\usepackage{subcaption}

\usepackage{amsmath, amssymb}
\usepackage{booktabs,tabularx}
\newcolumntype{Y}{>{\raggedright\arraybackslash}X}

\usepackage{multirow}
\usepackage{makecell}
\usepackage{wrapfig} 
\usepackage{caption}
\usepackage{microtype}
\usepackage{placeins}   

\usepackage{pifont}

\makeatletter
\newif\ifinappendix
\AddToHook{cmd/subsection/before}{\ifinappendix \FloatBarrier\clearpage \fi}
\AddToHook{cmd/subsubsection/before}{\ifinappendix \FloatBarrier \fi}
\makeatother

\title{Untangling Component Imbalance in Hybrid Linear Attention Conversion Methods}

\author{Martin Benfeghoul$^{1,}$\thanks{Equal Contribution.} , Teresa Delgado$^{1,*}$, Adnan Oomerjee$^{1, 2}$, Haitham Bou-Ammar$^{1,2}$, \\ \textbf{Jun Wang}$^{2}$, \textbf{Zafeirios Fountas}$^{1}$ \\
$^{1}$Huawei, Noah's Ark Lab, London \\
$^{2}$AI Centre, Department of Computer Science, University College London, London, UK\\
\texttt{\{martin.antoine.benfeghoul,teresa.delgado\}@h-partners.com} \\
\texttt{\{adnan.ebrahim.oomerjee,haitham.ammar\}@huawei.com} \\
\texttt{zafeirios.fountas@huawei.com} \\
\texttt{jun.wang@ucl.ac.uk}
}

\iclrfinalcopy 
\begin{document}

\maketitle

\begin{abstract}

Transformers' quadratic computational complexity limits their scalability despite remarkable performance. While linear attention reduces this to linear complexity, pre-training such models from scratch remains, in most cases, prohibitively expensive. Recent post-training linearisation methods convert pre-trained Transformers to linear models efficiently, often using hybrid approaches that combine linear attention with sliding-window softmax. We identify a critical flaw: existing hybrid methods inadvertently bypass the linear component, relying almost entirely on SWA. Component-level diagnostics reveal this previously undetected behaviour stems from overlooked evaluation practices on common-sense benchmarks. We propose three solutions to ensure balanced component usage: (i) inference-time hybridisation of linear-only conversions with sliding-window softmax; (ii) HedgeCATs, combining attention-weight transfer with targeted LoRA fine-tuning; and (iii) Scheduled Sliding-window Dropout (SSD), which stochastically suppresses the softmax branch during training to prevent component collapse. Our methods maintain computational efficiency while recovering most base model performance and ensuring genuine linear attention adoption, restoring the validity of performance attributions in hybrid conversions.
\end{abstract}

%
%
%
\section{Introduction}\label{sec:introduction}

Transformers \citep{vaswani2017attention} have delivered state-of-the-art results across language, vision, and multimodal tasks, yet their quadratic attention cost in sequence length remains a central bottleneck for long-context inference and training. Linear attention (LA) \citep{katharopoulos2020linearattention}, offers a compelling alternative by replacing the softmax kernel with linear feature maps that enable associative, streaming updates of a fixed-size recurrent state \citep{ choromanski2021rethinking, pmlr-v201-duman-keles23a, Banerjee2020ExploringAT, peng2021random, zhen2022cosformer}. In principle, this reduces the asymptotic complexity of both memory and compute compared to softmax attention. In practice, however, fully pre-training LA models is costly \citep{liu-etal-2020-understanding}, and performance often lags behind quadratic baselines trained with similar budgets due to limitations in representational complexity \citep{zhang2024hedgehog, mercat2024supra}.

A growing body of work focuses on circumventing the high cost of pre-training linear models via post-training linearisation \citep{zhang2024hedgehog, zhang2024lolcats, lan2025liger, mercat2024supra}: converting a pre-trained quadratic Transformer into a fully linear or hybrid linear-softmax model. This approach amortises most of the cost into the pre-trained base model and performs a light “swap + adaptation” stage, in which softmax kernel is replaced with a learnable linear kernel, with the model then put through additional pre-training and/or supervised fine-tuning to recover performance. Such methods have been reported to require on the order of $0.02\%$ (or less) of the data used to train the base model to recover performance. Existing conversion methods typically differ in (i) the LA formulation, (ii) whether and how the original weights are fine-tuned, and (iii) whether sliding-window softmax attention (SWA) \citep{beltagy2020longformer, NEURIPS2020_c8512d14} is retained alongside the linear path, yielding hybrid architectures.

Hybrid conversions are attractive as they pragmatically combine the representational capacity of SWA with the memory and computational efficiency of LA. This makes them powerful for long-context tasks. Yet, as we show, these reported gains can mask a critical failure mode: the model may lean almost entirely on the SWA path while effectively ignoring the linear component. This creates misleading performance attribution: the hybrid is credited for “using LA” when, in reality, the model simply biases and adapts itself towards SWA entirely during the adaptation stage. The field lacks standard diagnostics to quantify each component’s contribution, so such imbalances can remain hidden behind aggregate metrics.

\paragraph{Contributions} (1) We identify and characterise a systematic issue in current hybrid attention conversions whereby models learn to ignore their LA component and over-rely on their SWA one, leading to misleading attribution of hybrid performance. (2) We provide component-level diagnostic that make this imbalance visible and reproducible across popular pre-trained base models on standard common-sense benchmarks. (3) We introduce three practical remedies: a zero-shot inference-time hybrid; HedgeCATs, which combines HedgeHog-style attention-weight transfer with brief LoRA fine-tuning; and Scheduled Sliding-window Dropout (SSD) to prevent component imbalance during training. We show that our proposed strategies recover most base-model performance while ensuring genuine use of the LA pathway, restoring attributional validity without sacrificing computational efficiency.

\section{Background \& Related Works}\label{sec:related_works}
\subsection{Linear Attention}
Let $\mathbf{X}\!\in\!\mathbb{R}^{T\times d_{\text{model}}}$ be a sequence of length $T$ with projections
$\mathbf{Q}=\mathbf{X}\mathbf{W}_Q$, $\mathbf{K}=\mathbf{X}\mathbf{W}_K$, $\mathbf{V}=\mathbf{X}\mathbf{W}_V$, where $\mathbf{Q},\mathbf{K}\!\in\!\mathbb{R}^{T\times d}$ and $\mathbf{V}\!\in\!\mathbb{R}^{T\times d_v}$.
Standard softmax attention \citep{vaswani2017attention} computes
\begin{equation}
    \mathbf{O}=\mathrm{softmax}\!\left(\frac{\mathbf{Q}\mathbf{K}^\top}{\sqrt d}\right)\mathbf{V}    
\end{equation}
incurring $O(T^2)$ time and memory for the $T{\times}T$ similarity matrix. Using a kernel $\kappa(\mathbf{q},\mathbf{k})$ such that $\kappa(\mathbf{q},\mathbf{k})\approx \boldsymbol{\phi}(\mathbf{q})^\top \boldsymbol{\phi}(\mathbf{k})$ for a non-negative feature map $\boldsymbol{\phi}:\mathbb{R}^d\!\to\!\mathbb{R}^{d'}$ as proposed by~\cite{katharopoulos2020linearattention}, we can avoid forming pairwise similarities by introducing global summaries:
\begin{equation}
    \mathbf{o}_t=\frac{\sum_{i=1}^T \kappa(\mathbf{q}_t,\mathbf{k}_i)\,\mathbf{v}_i}{\sum_{i=1}^T \kappa(\mathbf{q}_t,\mathbf{k}_i)}
    =\frac{\boldsymbol{\phi}(\mathbf{q}_t)^\top \mathbf{S}}{\boldsymbol{\phi}(\mathbf{q}_t)^\top \mathbf{z}},\quad
    \mathbf{S}=\sum_{i=1}^T \boldsymbol{\phi}(\mathbf{k}_i)\mathbf{v}_i^\top,\ \
    \mathbf{z}=\sum_{i=1}^T \boldsymbol{\phi}(\mathbf{k}_i).
\end{equation}
These summaries can be accumulated in a single pass, so no $T{\times}T$ score matrix is materialised. This yields $O(Tdd')$ time and $O(dd')$ memory. 
Proposed $\boldsymbol{\phi}$ in the literature include non-negative element-wise maps (such as $1 + \mathrm{ELU}(\mathbf{x})$ \citep{katharopoulos2020linearattention} and $\mathrm{ReLU}$ \citep{Kasai2021T2R}), random feature maps \citep{choromanski2021rethinking}, and exponential function approximations via low-order Taylor expansions \citep{pmlr-v201-duman-keles23a, Banerjee2020ExploringAT}.

\subsection{Hybrid Attention}

Hybrid attention mechanisms combine softmax attention with LA through various architectural approaches. Some methods interleave full softmax attention layers with LA layers~\citep{lieber2024jamba,dong2024hymba}, while others employ SWA combined with LA, either in interleaved layers~\citep{ren2024samba} or integrated within the same Transformer block~\citep{beltagy2020longformer,zhang2024lolcats,lan2025liger,irie2025blending, munkhdalai2024leave}. The adoption of SWA offers the key advantage of preserving linear complexity throughout the entire model.
Following other hybrid conversion approaches~\citep{zhang2024lolcats,lan2025liger,irie2025blending}, our work focuses on combining SWA with LA within Transformer blocks to approximate full softmax attention. These methods typically employ a scaled linear combination of the two attention outputs, utilising learned~\citep{munkhdalai2024leave, zhang2024lolcats}, fixed~\citep{lan2025liger}, or data-dependent scaling factors~\citep{behrouz2024titans, irie2025blending}. This can be summarised by scalar or vector mixing terms $a,b$, such that the hybrid attention output is given by:
\begin{equation}
    \textbf{ATTN}(\mQ,\mK,\mV, \va, \vb) 
    = \va \odot \textbf{SWA}(\mQ,\mK,\mV) 
    + \vb \odot \textbf{LA}(\mQ,\mK,\mV)
\end{equation}\label{eq:hybrid_attention}

\subsection{Linearising Pre-Trained Transformers}

Although numerous linear~\citep{schlag2021deltanet, gu2023mamba, yang2023gla, yang2024gateddeltanet,peng2025rwkv} and hybrid~\citep{beltagy2020longformer, zhu2021longshort,lieber2024jamba,yang2024gateddeltanet,behrouz2024titans} Transformers have been developed, the majority are trained from scratch. The prohibitive cost of pre-training constrains most of these approaches to small model sizes (typically $<1$B parameters) and makes them expensive to reproduce or scale when checkpoints are unavailable, directly undermining the central promise of linear Transformers: computational efficiency. This limitation renders post-training conversion methods particularly appealing, as they offer the potential for high performance recovery in large-scale Transformers at a fraction of pre-training costs (typically $<1$\%).

While LA is theoretically equivalent to softmax-based self-attention under feature map $\phi$ and kernel $\phi(\vq_t)^\top \phi(\vk_i) = \text{exp}\left((\vq_t^\top \vk_i)\cdot D^{-1/2}\right)$~\citep{katharopoulos2020linearattention}, such a (finite-dimensional) feature map does not currently exist. As such, converting a pre-trained Transformer to use LA requires some adjustments and fine-tuning to make up for the change in attention weights.

Proposed kernels are designed to ensure positive attention weights via non-linear activation functions \citep{katharopoulos2020linearattention, Kasai2021T2R,mercat2024supra, zhang2024hedgehog, zhang2024lolcats}. However, these non-negative activation functions run the risk of suppressing any negative signals and may unnecessarily constrain the learned mappings. To this end,~\cite{zhang2024hedgehog} and~\cite{zhang2024lolcats} concatenate the negative mapping to the output along the head dimension, applying their respective non-linear activation function $\sigma$ to each separately:
\begin{equation}\label{eq:feature_map}
    \phi(\vx) = [ \sigma ( \mW^{\top}_{\phi}\vx + \vb ) \oplus \sigma ( -\mW^{\top}_{\phi}\vx - \vb ) ]
\end{equation}
\citet{zhang2024hedgehog} showed that the softmax function's unique spikiness and monotonicity with respect to the Query-Key dot-product are hard to match when using previously proposed candidates for $\phi$. They therefore learn an exponential feature map (Eq.~\ref{eq:feature_map} with $\sigma=\text{exp}(\cdot)$), and train it via an attention-weights transfer objective that minimises cross-entropy between softmax attention weights and linear weights. This ``weights-to-weights" stage is followed by a fine-tuning stage of the original model weights which they claim makes up for any approximation errors. LoLCATs~\citep{zhang2024lolcats} then later explores general conversion methods using LoRA~\citep{hu2022lora} fine-tuning as well as hybrid attention methods. In parallel, SUPRA \citep{mercat2024supra} follows T2R \citep{Kasai2021T2R} in adopting a ReLU-activated feature map with standard language-model fine-tuning rather than an explicit attention-transfer loss. This keeps training simple since no new parameters are introduced.

Other conversion methods \citep{Mao2022finetuningPTdecay, lan2025liger} focus on converting pre-trained Transformers to gated linear/recurrent blocks (GLA-style). Both these methods repurpose the attention block into a gated linear update; Liger \citep{lan2025liger} differs in explicitly retaining a local softmax branch. A summary of different linear kernels and transfer objectives for various conversion methods can be found in Table~\ref{tab:linearising-conversions}.

\subsection{Training Interventions to Tackle Component Collapse}
As explored in this paper, hybrid attention conversion models can learn to ignore the linear path and rely solely on SWA. Related work in other settings tackles analogous ``path collapse” with structured dropout \citep{srivastava2014dropout}: dropping whole substructures during training so models learn to balance all their trained components. In Transformers, DropHead \citep{zhou-etal-2020-scheduled} targets multi-head attention \citep{cordonnier2021multihead} directly by stochastically dropping whole heads with a scheduled rate to prevent a few heads from monopolising computation. Other works explore similar ideas but applied to different subcomponents such as Transformer layers \citep{Fan2020Reducing}, experts \citep{ChenZJLW23} in Mixture-of-Experts models, and even incoming keys \citep{li2023dropkey}.

\section{Identifying the Issues within Hybrid Conversion Methods}\label{sec:initial_analysis}
In this section, we outline some issues we have found to occur in conversion methods which make use of a hybrid attention-based training objective. We start by re-implementing and ablating the LoLCATs framework, which is considered to be the state-of-the-art (SOTA) method for converting pre-trained Transformers to use hybrid attention, as well as repeating such analysis in their own codebase and checkpoints. We complete these findings with a component-wise investigation of similar methods, ablating key components with those used in similar SOTA LA-only methods, one at a time, in order to identify the ones responsible for the issues observed in hybrid methods.

\subsection{Experimental Set-up}\label{subsec:exp_setup}
\paragraph{Feature Map $\Phi$}
We adopt a learned feature map~\citep{mercat2024supra, zhang2024hedgehog, zhang2024lolcats}, with $\mW_\phi \in \mathbb{R}^{h_d \times \frac{h_d}{2}}$ and a softmax activation function (ie. Eq~\ref{eq:feature_map} with $\sigma= \text{softmax}(\cdot)$), and apply the RoPE embeddings to queries and keys prior to applying $\phi$, as motivated by~\cite{zhang2024lolcats}.

\paragraph{Combining Sliding-Window and Linear Attention} For our experiments, we implement Eq.~\ref{eq:hybrid_attention} with a simple choice of $\va=\vg$, $\vb=\mathbf{1}-\vg$, with fixed $\vg=\tfrac{1}{2}\mathbf{1}$, as used in~\cite{lan2025liger}. We avoid learned or dynamic mixing terms to ensure that LA and SWA are used equally in the hybrid attention outputs. The LA component only operates on tokens outside the sliding-window, which has size $64$.

\paragraph{Models and Datasets}
In our experiments, we convert and evaluate up to three popular, pre-trained Transformers: Mistral-7B-Instruct-v0.1, Llama3-8B-Instruct, and Llama3.1-8B-Instruct\footnote{Note that we remove the "-Instruct" from the model names in our results for brevity but still refer to these models rather than the base models.}. Attention transfer and fine-tuning are carried out on truncated samples of 1024 tokens from the FineWeb-Edu dataset~\citep{penedo2024fineweb}. For evaluation, we report performance on popular LM-Eval tasks, all zero-shot: PIQA, ARC-Easy, ARC-Challenge (acc norm), HellaSwag (acc norm), WinoGrande, and MMLU. In all results, any average performance shown is calculated across all six tasks.

\paragraph{Training \& LoRA Parameters} We follow~\cite{zhang2024lolcats}'s exact fine-tuning settings, applying LoRA to the query, key, value, and output projection matrices ($\mW_\vq$, $\mW_\vk$, $\mW_\vv$, $\mW_\vo$), with rank $r=8$, $\alpha=16$. We use an AdamW~\citep{loshchilov2017adamw} optimiser with learning rate $1e-4$ ($1e-2$ during attention transfer), and a reduce-on-plateau scheduler. We train for $1$ epoch for attention transfer, and 1-5 epochs during fine-tuning, using an effective (accumulated) batch size of 64 and approximately $25$M tokens per epoch.

\subsection{Ablation Study of LoLCATs Hybrid Attention Modules at Inference time}\label{results:initial_findings}
In this section, we explore the contribution of each attention module to performance for models trained using LoLCATs' proposed hybrid conversion methodology~\citep{zhang2024lolcats}. We run four ablations at inference time: (i) SWA-only, disabling the LA module by forcing its outputs to zeros; (ii) LA-only where we disable SWA and retain only the LA component; (iii) attention sinks~\citep{xiao2023attnsinks} only, suppressing both SWA and LA and only passing the first $8$ values through softmax attention; and (iv) no attention where we return an all-zeros attention output, removing any contribution from the attention mechanism. Additionally, we run the same ablations using the provided LoLCATs-trained checkpoints for Llama-3.1-8B using the LoLCATs codebase. Results are shown in Table~\ref{lolcats-hybrid-ablation}.

Together, the ablations show that almost all the performance attribution sits within the SWA component. Using SWA-only in the resulting models either matches or slightly improves average accuracy for both Mistral-7B and the LoLCATs-trained Llama-3.1-8B \footnote{\url{https://huggingface.co/hazyresearch/lolcats-Llama-3.1-8b-distill} \url{https://huggingface.co/hazyresearch/lolcats-Llama-3.1-8b-ft-lora}}. By contrast, LA-only, attention sinks only, and no attention all collapse to roughly the same low performance. The results expose a clear SWA-LA imbalance in current hybrid attention conversion methods that leads to LA contributing little to downstream accuracy and even being detrimental. Similar findings are reported in \cite{lan2025liger} (see their Table 6) where they find that SWA-only and GLA+SWA lead to very similar performance while GLA-only leads to a dramatic decrease in performance.

\begin{table}[ht!]
\centering
\resizebox{\linewidth}{!}{
\begin{tabular}{l | c c c c c c | c | c}
\toprule
\textbf{Active Attn Modules} & \textbf{PIQA} & \textbf{ARC-E} & \textbf{ARC-C} & \textbf{HellaSwag} & \textbf{WG} & \textbf{MMLU} & \textbf{AVG} & \textbf{Rec. Perf} \\
\midrule
    \textit{Mistral-7B} & 79.27 & 80.01 & 52.22 & 74.60 & 69.93 & 53.51 & 68.26 & 100.00 \\
    \quad SWA + Linear & 78.40 & \textbf{79.50} & \textbf{49.91} & \textbf{71.28} & \textbf{68.35} & 45.44 & 65.48 & 95.93 \\
    \quad SWA only  &  \textbf{78.56} & 79.46 & 49.83 & 70.57 & \textbf{68.35} & \textbf{46.56} & \textbf{65.56} & \textbf{96.04}  \\
    \quad Linear only &  53.65 & 29.71 & 24.57 & 27.01 & 50.43 & 23.28 & 34.78 & 50.95\\
    \quad Attn sinks only & 54.41 & 30.05 & 24.66 & 28.80 & 49.64 & 23.67 & 35.21 & 51.58 \\
    \quad No Attention & 53.92 &  25.63 & 24.66 & 25.99 & 50.67 & 25.51 & 34.40 & 50.39 \\
\midrule
    \textit{Llama3-8B} & 78.13 & 81.69 & 56.66 & 75.94 & 71.67 & 63.85 & 71.32 & 100.00 \\
    \quad SWA + Linear & \textbf{78.35} & \textbf{80.47} & \textbf{54.35} & \textbf{71.62} & \textbf{71.67} & \textbf{48.84} & \textbf{67.55} & \textbf{94.71} \\
    \quad SWA only  & 78.02 & 80.01 & 54.10 & 65.08 & 71.98 & 47.30 & 66.08 & 92.26 \\
    \quad Linear only & 52.07 & 25.97 & 25.77 & 26.31 & 48.22 & 22.95 & 33.55 & 47.04 \\
    \quad Attn sinks only & 56.31 & 29.00 & 22.61 & 28.83 & 49.41 & 22.95 & 34.85 & 48.86 \\
    \quad No Attention & 55.17 & 26.73 & 22.87 & 26.13 & 51.14 & 22.95 & 34.17 & 47.91 \\
\midrule
    \textit{Llama3.1-8B (LoLCATs ckpt)} & 80.14 & 81.82 & 55.20 & 79.14 & 73.72 & 68.05 & 73.01 & 100.00 \\
    \quad SWA + Linear & 81.18 & \textbf{82.37} & 54.78 & 79.16 & 70.09 & \textbf{58.89} & 71.08 & 97  \\
    \quad SWA only  & \textbf{81.56} & \textbf{82.37} & \textbf{55.29} & \textbf{79.76} & \textbf{74.11} & 55.63 & \textbf{71.45} & \textbf{98}   \\
    \quad Linear only & 51.52 & 25.00 & 25.51 & 26.37 & 52.25 & 23.09 & 33.96 & 47 \\
    \quad Attn sinks only & - & - & - & - & - & - & - & -  \\
    \quad No Attention & 54.62 & 26.68 & 24.40 & 25.88 & 48.86 & 23.12 & 33.93 & 46 \\

\bottomrule
\end{tabular}
}
\caption{Measuring the effect of attention components on benchmark accuracy for models trained with LoLCATs hybrid conversion.}
\label{lolcats-hybrid-ablation}
\end{table}

\subsection{Reverting Back to Linear Attention-only}\label{results:HedgeHog}
In this section, we re-examine the HedgeHog~\citep{zhang2024hedgehog} conversion method with LoLCATs'~\citep{zhang2024lolcats} fine-tuning settings. HedgeHog has been shown to work with full-parameter fine-tuning and may be considered the SOTA conversion method for LA-only. Hence, we simply seek to determine whether a more conservative, LoRA-based fine-tune is also able to make use of LA-only in this setting, or whether this difference may be responsible for the gap in LA-only performance between HedgeHog and LoLCATs. More specifically, we evaluate models converted using HedgeHog's core methods, namely attention transfer on attention weights using soft-label cross-entropy and a square projections matrix $\mW_\phi \in \mathbb{R}^{h_d \times h_d}$, but change the fine-tuning stage to use LoLCATs' LoRA settings. Our findings are illustrated in Figure~\ref{fig:la_only}, while per-task results are provided in Appendix~\ref{appdx:linear_only}.

\begin{figure}[h!]
    \centering
    \includegraphics[width=0.9\linewidth]{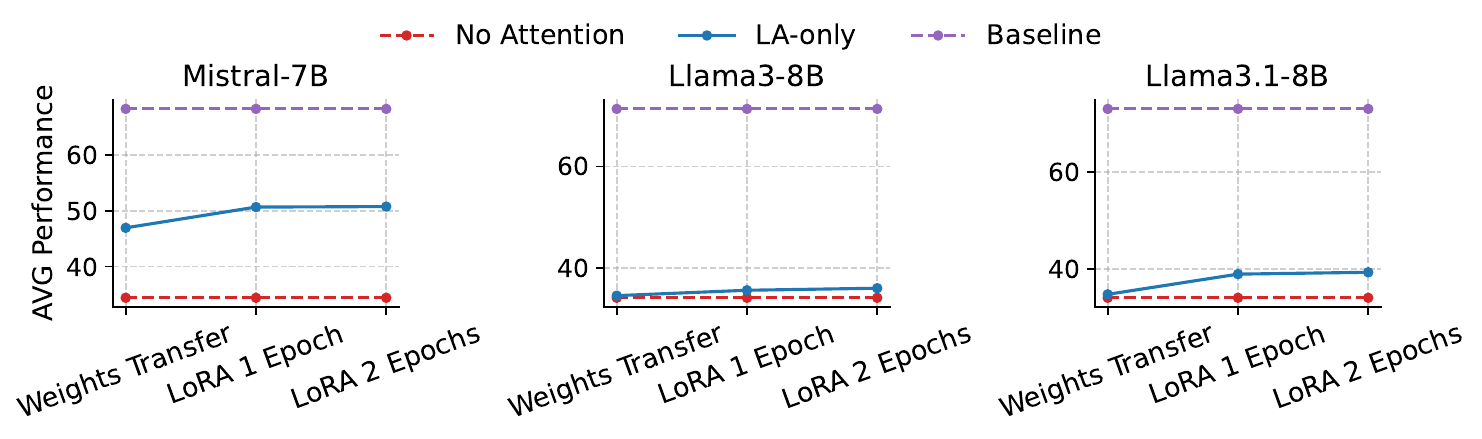}
    \caption{Measuring the performance of HedgeHog conversion at various points during the conversion process. LoRA refers to a LA-only fine-tuning.}
    \label{fig:la_only}
\end{figure}

We observe a clear contrast with the results in section~\ref{results:initial_findings}, with the linear component in all models beating the no-attention ablation with no or little fine-tuning. However, we observe that the Llama models, and Llama3-8B especially, only see a minor improvement in performance over no use of attention at all. 

\subsection{Component-Wise Investigation}\label{results:component_wise_investigation}
Our minimal implementation of the HedgeHog conversion method successfully makes use of LA, although with mixed performance recovery. On the other hand, LoLCATs claims to recover base model performance, but appears to do so entirely using SWA. In this section, we seek to identify which parts of these two methods contribute to these two different results. To this end, we ablate various parts of these methods within the attention transfer stage and evaluate them with LA-only. Focusing on this stage of the conversion process allows us to determine exactly which components work best with LA without affecting any weights shared with SWA during fine-tuning. We focus our ablations on the Mistral-7B model as it saw the greatest difference in LA performance when changing the attention transfer and projections $\phi$ from LoLCATs' to HedgeHog's.

\subsubsection{Attention Transfer Learning}\label{results:transfer_objective}
In this ablation, we keep the square HedgeHog feature map used in section~\ref{results:HedgeHog} and simply ablate the Attention Transfer objective. We evaluate three modes: (1) The HedgeHog attention weights transfer using soft-label cross entropy loss between softmax attention weights and the Hedgehog attention weights (both quadratic cost) (2) The LoLCATs hybrid attention outputs transfer using MSE loss between full-context softmax attention outputs (quadratic cost) and the hybrid SWA + LA outputs (linear cost), and (3) the MSE between full softmax attention outputs (quadratic cost) and the full LA outputs (linear cost). We omit any LoRA fine-tuning in these evaluations to ensure we measure the direct impact of attention transfer only. Our findings are outlined in Table~\ref{tab:TransferObjective}, while our results for all models are included in Appendix~\ref{appdx:results_transfer_objective}.

\begin{table}[h!]
\centering
\resizebox{\linewidth}{!}{
\begin{tabular}{l | c c c c c c | c | c}
\toprule
\textbf{Transfer Objective} & \textbf{PIQA} & \textbf{ARC-E} & \textbf{ARC-C} & \textbf{HellaSwag} & \textbf{WG} & \textbf{MMLU} & \textbf{AVG} & \textbf{Rec. Perf} \\
\midrule
    \textit{Mistral-7B} & 79.27 & 80.01 & 52.22 & 74.60 & 69.93 & 53.51 & 68.26 & 100.00 \\
    \quad Attention Weights & 67.90 & 61.74 & 29.69 & 45.82 & \textbf{52.64} & 23.82 & 46.94 & 68.76 \\
    \quad Attention Outputs & \textbf{69.64} & \textbf{62.46} & \textbf{31.40} & \textbf{46.61} & 50.28 & 23.00 & \textbf{47.23} & \textbf{69.20} \\
    \quad Hybrid Attention Out. & 55.44 & 32.83 & 24.06 & 27.36 & 49.80 & 22.98 & 35.41 & 51.88 \\
    \quad No Attention & 53.92 & 25.63 & 24.66 & 25.99 & 50.67 & \textbf{25.51} & 34.40 & 50.39 \\
\bottomrule
\end{tabular}
}
\caption{Comparing the performance of LA-only after the attention transfer phase across learning objectives used during this stage.}\label{tab:TransferObjective}
\end{table}

Our results suggest that the hybrid attention output objective is likely to be responsible for LoLCATs failure to make use of LA, as it barely beats no attention at all. On the other hand, using a full attention output objective seems to beat even the weights transfer objective in this setting. In fact, this objective showed a particular advantage with Llama3-8B, although both Llama models still lag behind Mistral's recovered performance. We note that we have not observed any successful LA-only conversions of these models in the literature, as opposed to Mistral, suggesting that they may be particularly hard to convert. We leave the analysis as to why this might be to future work.

\subsubsection{\texorpdfstring{$\Phi$}{Phi} Dimensionality}\label{results:phi_dimensionality}
We now seek to measure the impact of the feature map size on conversion success. The original HedgeHog paper makes use of a square linear map $\mW_\phi \in \mathbb{R}^{h_d \times h_d}$ ($\phi: \mathbb{R}^{h_d} \rightarrow \mathbb{R}^{2h_d}$, given Eq.~\ref{eq:feature_map}), with $h_d$: query-key head dimension, but LoLCATs reduce this to a rectangular linear map $\mW_\phi \in \mathbb{R}^{h_d \times \frac{h_d}{2}}$, for a significantly smaller $\phi: \mathbb{R}^{h_d} \rightarrow \mathbb{R}^{h_d}$. To this end, we repeat the experiments presented above in Table~\ref{tab:TransferObjective}, with $\mW_\phi \in \mathbb{R}^{h_d \times \frac{h_d}{2}}$. Our findings are outlined in Table~\ref{tab:feature_map_size}.

\begin{table}[h!]
\centering
\resizebox{\linewidth}{!}{
\begin{tabular}{l | c c c c c c | c | c}
\toprule
\textbf{Transfer Objective} & \textbf{PIQA} & \textbf{ARC-E} & \textbf{ARC-C} & \textbf{HellaSwag} & \textbf{WG} & \textbf{MMLU} & \textbf{AVG} & \textbf{Rec. Perf} \\
\midrule
    \textit{Mistral-7B} & 79.27 & 80.01 & 52.22 & 74.60 & 69.93 & 53.51 & 68.26 & 100.00 \\
    \quad Attention Weights & \textbf{64.85} & \textbf{54.84} & 26.11 & \textbf{39.48} & \textbf{51.07} & 23.17 & \textbf{43.25} & \textbf{63.37} \\
    \quad Attention Outputs & 63.71 & 53.83 & \textbf{27.30} & 38.50 & 49.33 & 22.90 & 42.60 & 62.40 \\
    \quad Hybrid Attention Out. & 53.97 & 30.81 & 22.44 & 27.42 & 49.88 & 22.99 & 34.59 & 50.67 \\
    \quad No Attention & 53.92 & 25.63 & 24.66 & 25.99 & 50.67 & \textbf{25.51} & 34.40 & 50.39 \\
\bottomrule
\end{tabular}
}
\caption{Repeating ablations described in Table~\ref{tab:TransferObjective}, but with half the $\mW_\phi$ size.}\label{tab:feature_map_size}
\end{table}

Our results broadly follow that of section~\ref{results:transfer_objective}, with performance decreasing across the board with this reduction of output features. More specifically, we see the hybrid attention objective still improves performance only marginally, while the attention weights and outputs objectives achieve a significant and similar improvement on no attention. We note that the weights transfer objective appears to have the upper hand with this smaller $\mW_\phi$. 

\subsubsection{\texorpdfstring{$\Phi$}{Phi} Activation Function}
Finally, HedgeHog implements an exponential activation function ($\sigma = \text{exp}(\cdot)$). On the other hand, LoLCATs use the softmax, which is the layer-normalised equivalent ($\sigma = \text{softmax}(\cdot)$). Here, we ablate the two, and compare the resulting performance along with T2R and SUPRA's ReLU activation, as well as no activation function at all. Our findings are outlined in Table~\ref{tab:feature_map_activation_fn}.

\begin{table}[h!]
\centering
\resizebox{\linewidth}{!}{
\begin{tabular}{l | c c c c c c | c | c}
\toprule
\textbf{\texorpdfstring{$\Phi$}{Phi} Activation Fn} & \textbf{PIQA} & \textbf{ARC-E} & \textbf{ARC-C} & \textbf{HellaSwag} & \textbf{WG} & \textbf{MMLU} & \textbf{AVG} & \textbf{Rec. Perf} \\
\midrule
    \textit{Mistral-7B} & 79.27 & 80.01 & 52.22 & 74.60 & 69.93 & 53.51 & 68.26 & 100.00 \\
    \quad Softmax & \textbf{67.90} & \textbf{61.74} & \textbf{29.69} & \textbf{45.82} & \textbf{52.64} & 23.82 & \textbf{46.94} & \textbf{68.76} \\
    \quad Exponential & 66.32 & 61.45 & \textbf{29.69} & 44.67 & 50.99 & 23.52 & 46.11 & 67.55 \\
    \quad ReLU & 56.53 & 34.81 & 22.95 & 27.95 & 50.91 & 23.29 & 36.07 & 52.85 \\
    \quad $1 +$ ELU & 54.68 & 29.12 & 23.81 & 25.98 & 51.46 & \textbf{25.64} & 35.12 & 51.45 \\
    \quad None & 51.74 & 25.51 & 29.61 & 25.97 & 51.07 & 24.95 & 34.81 & 51.00 \\
    \quad No Attention & 53.92 & 25.63 & 24.66 & 25.99 & 50.67 & 25.51 & 34.40 & 50.39 \\
\bottomrule
\end{tabular}
}
\caption{Ablating the activation function used in HedgeHog projections ($\phi$). All results show checkpoints for a single epoch of weights transfer with no fine-tuning, evaluated as LA-only.}\label{tab:feature_map_activation_fn}
\end{table}

Our results show the the exponential-based activation functions (exponential and softmax) far outperform the alternatives. In this regard, LoLCATs and HedgeHog appear to achieve comparable performance, with a slight edge in LoLCATs. Interestingly, $1+$ELU does not appear to carry this same performance despite its close similarity to the exponential function by itself.

\section{Methods \& Results}\label{sec:results}
Section~\ref{sec:initial_analysis} demonstrates some key issues in hybrid conversion methods, and isolates the components responsible for such shortcomings, namely the hybrid attention transfer objective and the dimensionality of $\phi$. 
In this section, we build on top of these findings and propose, evaluate, and compare three different methods to fine-tune a model for hybrid attention while avoiding the dominance of SWA and decay of LA. Note that all experiments conducted in this section use the same experimental setup as in Subsection~\ref{subsec:exp_setup}, unless explicitly stated otherwise.

\subsection{Inference-Time Hybrid}
Seeing as fine-tuning for hybrid attention appears to encourage the model to focus on the more expressive SWA, we first propose a ``zero-shot" hybrid, introducing SWA in models which have either seen no fine-tuning or fine-tuning with LA-only. We implement two modes of hybrid attention: one where LA only sees tokens outside of SWA's sliding window, and one where LA sees all past tokens therefore overlapping with SWA's context ("Overlap" in our results). Our findings are illustrated in Figure~\ref{fig:inference_time_hybrid}, while per-task results are provided in Appendix~\ref{appdx:results_inference_time_hybrid}.

\begin{figure}[h!]
    \centering
    \includegraphics[width=0.9\linewidth]{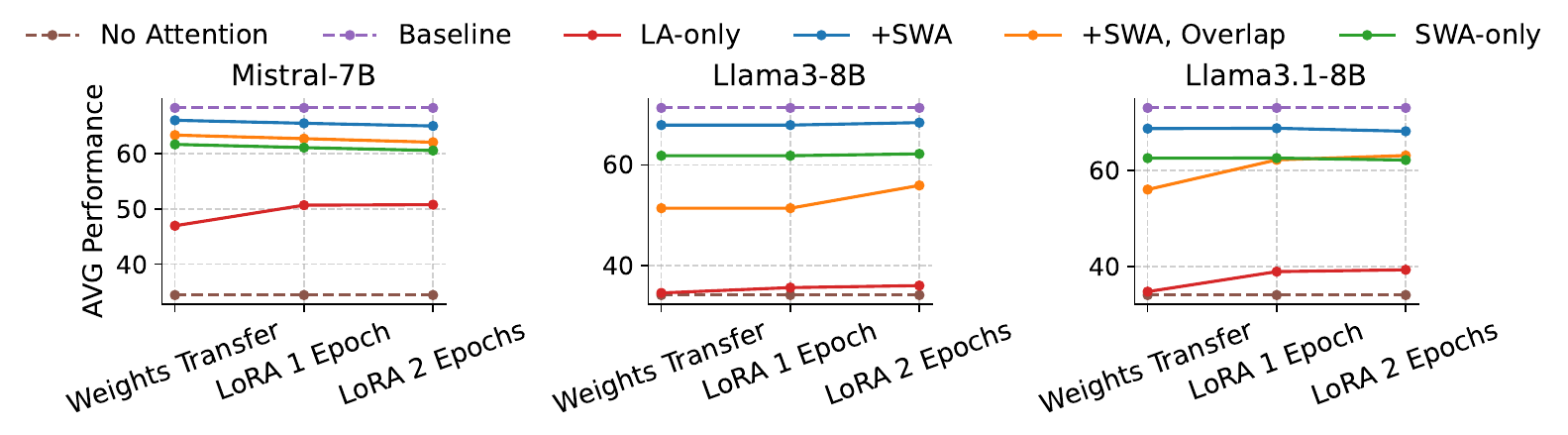}
    \caption{Adding SWA at inference time at various points of the conversion process.}
    \label{fig:inference_time_hybrid}
\end{figure}

We find that, while LA-only benefits from fine-tuning, the latter appears to slightly degrade the performance of SWA. As such, the best inference-time hybrid performance is achieved with no fine-tuning. Furthermore, it would appear as though overlapping LA's context with that of SWA also degrades performance. We theorise that this is most likely due to our fixed mixing term $g=0.5$ and may be improved by a data-dependent mixing term which may dynamically choose how to weight the contributions of each component in the case of overlap, as employed by~\cite{irie2025blending}. 

\subsection{HedgeCATs: HedgeHog Transfer + LoLCATs fine-tuning}
Our second method, HedgeCATs, blends HedgeHog's attention weight transfer with LoLCATs' LoRA fine-tuning of hybrid attention. The first training stage performs HedgeHog-style transfer to learn a feature map $\phi$ that mimics full softmax attention, training with LA-only so the LA path stands up on its own.
\begin{wrapfigure}{r}{0.5\textwidth}
  \vspace{-\baselineskip}
  \centering
  \includegraphics[width=\linewidth]{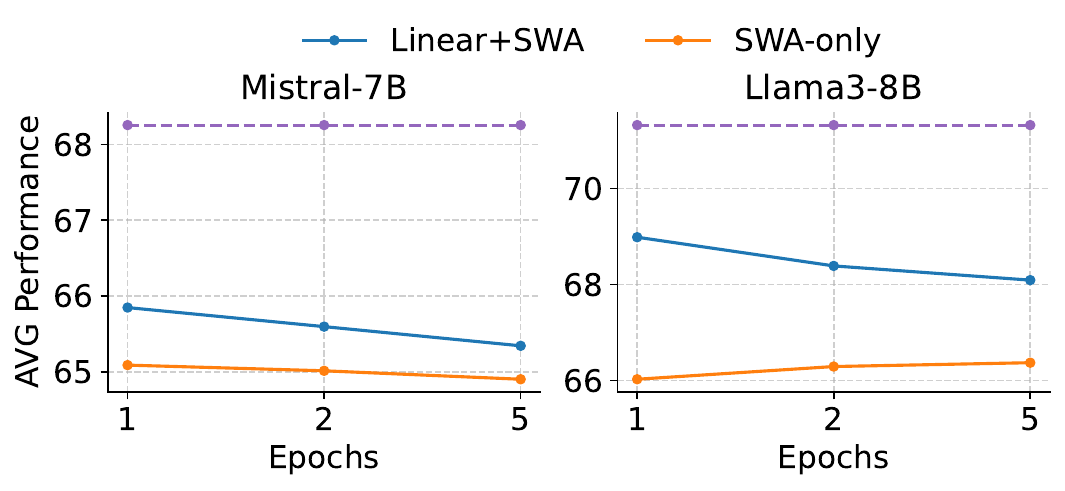}
  \captionsetup{aboveskip=2pt, belowskip=0pt}
  \caption{The average performance of HedgeCATs-trained models for difference amounts of LoRA fine-tuning.}
  \label{fig:hedgecats}
\end{wrapfigure}
Stage 2 applies LoRA fine-tuning while re-introducing SWA, aiming to recover base-model accuracy without letting the hybrid collapse back onto SWA. In practice, early stopping is key: Figure \ref{fig:hedgecats} shows that for Llama-3 8B, as fine-tuning proceeds the hybrid branch trends down, whereas SWA-only improves; for Mistral-7B, both hybrid and SWA-only degrade with more LoRA steps. These behaviours suggest a very short LoRA schedule gives the best trade-off between accuracy, attributional validity of hybrid attention, and training budget. Per-task performance can be seen in Appendix~\ref{appdx:results_hedgecats}.

\subsection{SSD: Scheduled Sliding-window Dropout}
\begin{figure}[tbp]
  \centering
  \begin{subfigure}{0.5\textwidth}
    \includegraphics[width=\linewidth]{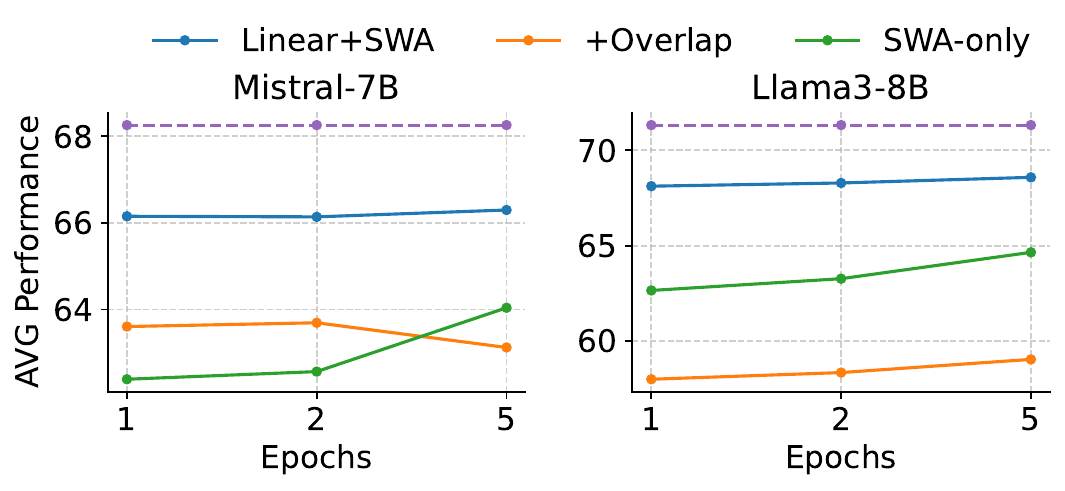}
    \caption{Scheduled dropout 0.9, 0.75, 0.5; fixed SW 32}
    \label{fig:schedules:a}
  \end{subfigure}\hfill
  \begin{subfigure}{0.5\textwidth}
    \includegraphics[width=\linewidth]{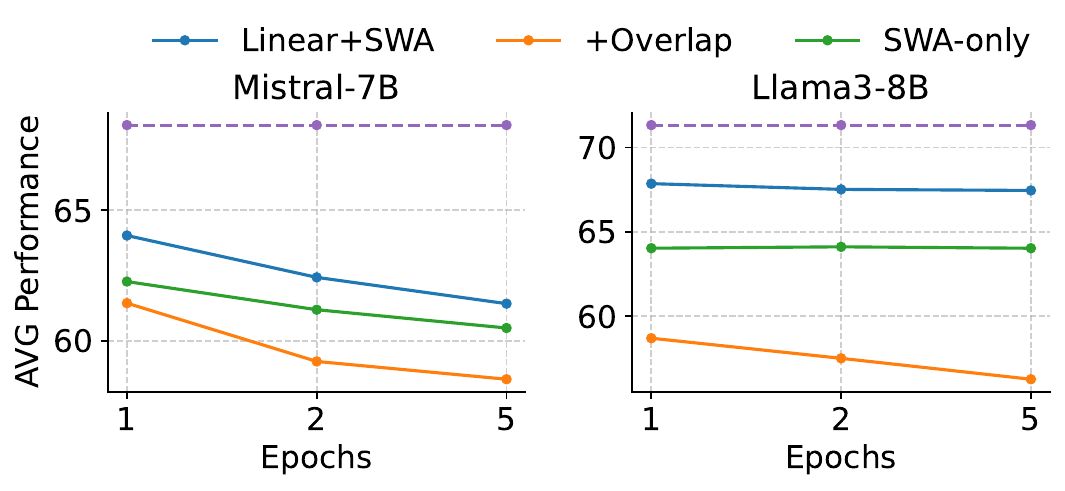}
    \caption{Fixed dropout 0.5; scheduled SW 4,8,16,32,64}
    \label{fig:schedules:b}
  \end{subfigure}

  \vspace{0.8em}

  \begin{subfigure}{0.5\textwidth}
    \includegraphics[width=\linewidth]{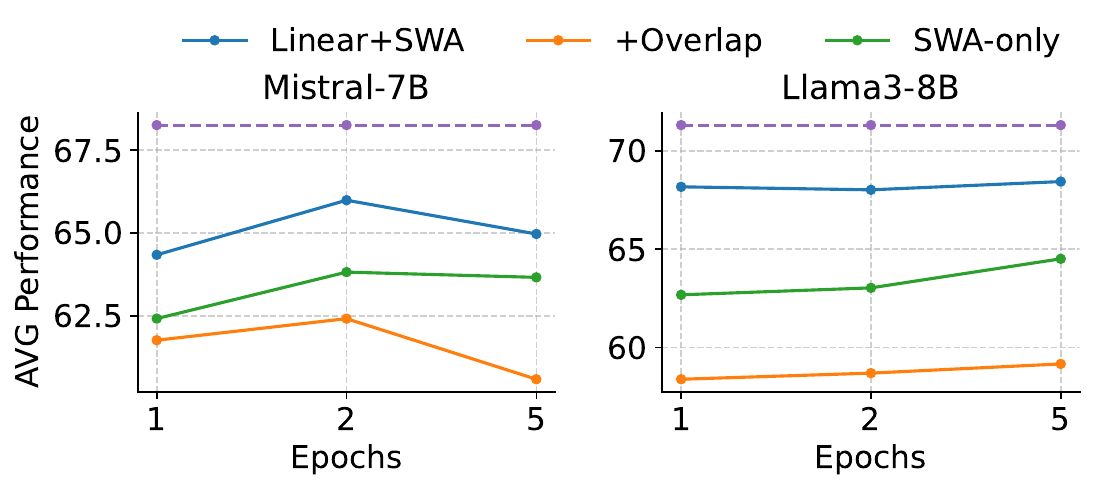}
    \caption{Dropout 0.5; SW 16}
    \label{fig:schedules:c}
  \end{subfigure}
  \caption{Performance comparison of dropout and sliding-window size schedules for a different number of fine-tuning epochs.}
  \label{fig:schedules}
\end{figure}

Finally, to further guard against SWA dominance during training, we introduce Scheduled Sliding-window Dropout (SSD). SSD alters the SWA component in hybrid attention during LoRA fine-tuning according to a dropout and sliding-window size schedule. Our results across SSD fine-tuning regimes are summarised in Figure~\ref{fig:schedules}. We vary dropout and sliding-window schedules (either fixed or epoch-varying) and evaluate at multiple fine-tuning epochs to characterise performance as a function of training time. Scheduled parameters are applied per epoch: at epoch \(k\) we use the \(k\)-th value in the schedule, and once the schedule is exhausted the final value is held fixed. This analysis illustrates how the SWA component of the hybrid attention mechanism evolves during fine-tuning. Figure~\ref{fig:schedules:a} uses a dropout schedule (0.9→0.75→0.5), i.e., the SWA branch is dropped 90\% of the time in epoch 1, with a fixed sliding-window size of 32; Figure~\ref{fig:schedules:b} fixes dropout at 0.5 and schedules the sliding-window size  (4→8→16→32→64); Figure~\ref{fig:schedules:c} fixes both dropout and sliding-window size at 0.5 and 16, respectively. Per-task results for each experimented SSD setting are provided in Appendix~\ref{appdx:results_ssd}.

The results for SSD-trained models show consistent trends across Mistral-7B and Llama-3-8B. For Figure~\ref{fig:schedules:a}, LA+SWA improves steadily with fine-tuning. This indicates that heavy early SWA dropout successfully pushes learning through the linear path before relaxing to 0.5. In Figure~\ref{fig:schedules:b}, performance starts lower compared to the previous experiment (penalised by the short initial window size) and remains flat or slightly down for Mistral-7B, with only a mild late recovery for Llama3-8B; SWA-only also drifts down or flattens, suggesting limited benefit from widening the window alone in this setting. The fixed dropout and sliding-window model, Figure~\ref{fig:schedules:c}, yields similar results to Figure~\ref{fig:schedules:a} for Llama and slightly worse performance for Mistral. Across experiment settings, SWA-only sits well below Linear+SWA, confirming that our fine-tuning schedule leads to component-balanced hybrid attention.

\section{Discussion \& Conclusion}\label{sec:discussion}
Our analysis reveals a critical failure mode in hybrid attention conversion methods: models trained with hybrid objectives often bypass LA entirely, relying exclusively on SWA. In contrast, attention weights or full attention outputs transfer objectives successfully enable LA, though with varying performance recovery.
We also confirm previous findings within our corrected framework, namely that exponential-family activations outperform alternatives, and larger feature map projections improve performance. These findings likely extend to other hybrid methods, like Liger, which report similar component imbalances.
To address this issue, we proposed three solutions for converting a model to use hybrid attention: (1) Inference-time hybrid addition, which preserves LA without additional training; (2) HedgeCATs, combining successful weights transfer with careful early-stopping during hybrid fine-tuning; and (3) Scheduled Sliding-window Dropout (SSD), providing robust training that maintains component balance. Each offers different trade-offs between simplicity, individual component performance, and training cost, while all recovering over $95\%$ of base model performance. Importantly, the SWA-driven LoLCATs-trained hybrids (Table~\ref{lolcats-hybrid-ablation}) consistently underperform compared to all our proposed methods. This indicates that enforcing genuine LA utilisation improves overall accuracy as well as attributional validity.

\paragraph{Limitations and Future Work}
In this study, performance is limited by our simplified implementation of some of the components. For example, it should be noted that replacing our fixed mixing term $g$ with learned and dynamic mixing mechanisms, as seen in other methods (see section~\ref{sec:related_works}) is likely to increase performance. Although we motivate such choices with this study's focus on clear performance attribution between LA and SWA, while minimising the model's ability to discard LA, future work should extend our resulting methods to maximise performance through further investigation of components such as the mixing term, normalisation methods, training datasets, LoRA settings, base models (instruction-tuned or not), etc.

LA is also believed to suffer from the same capacity issues observed in associative memory networks~\citep{schlag2021deltanet}. To this end, multiple gating mechanisms have been added to LA in order to manage information retention and retrieval accuracy across longer sequences~\citep{schlag2021deltanet, sun2023retnet, yang2023gla, yang2024gateddeltanet, liu2024longhorn}. Future work should examine how conversion methods may be applied to convert models to use such mechanisms. Additionally, LA conversion methods have the potential to improve performance of grouped KV retrieval methods~\citep{xiao2024infllm, fountas2025emllm}, which reduce information dilution in long-context settings by only attending to subsets of past tokens. Such methods are limited by the memory requirements for storing long token histories. 
An interesting extension of this work would use LA to compress individual memory blocks, reducing memory requirements while limiting the capacity issues arising from LA being applied to entire long-context sequences.

\paragraph{Conclusion}
In conclusion, while hybrid conversions promise efficiency with minimal performance loss, without careful design they fail to genuinely adopt LA. By identifying this failure mode and proposing solutions that maintain component balance, we restore attributional validity, ensuring claimed architectural components actually contribute to model behaviour, which is essential for advancing efficient Transformer architectures.

%
%
%

\subsubsection*{Acknowledgments}
The authors would like to thank Zachary Weller-Davies for his valuable input.

\subsection*{Ethics Statement}
This work focuses on foundational methods for converting large language models to use LA mechanisms. All experiments were conducted using publicly-available pre-trained models, and widely used training and benchmark datasets. No personally identifiable or sensitive data was used. The primary contribution is methodological, and thus does not introduce new societal risks beyond those already known for these models, such as potential biases or misuse. We encourage responsible and transparent use of these methods.

\subsection*{Reproducibility Statement}
We include experimental details across sections~\ref{sec:related_works}-~\ref{sec:results}, as well as Appendix~\ref{appdx:implementation_details}. Any details which aren't explicitly mentioned in this work are aligned with related works and clearly mentioned as such. Our experiments are reproducible within the LoLCATs public codebase with relatively minimal changes, and we are working on releasing our own public version of our codebase including all mentioned methods and ablations.

\bibliography{bibliography}
\bibliographystyle{iclr2026_conference}

\newpage
\appendix
\section{Appendix}
\subsection{Notation}

\paragraph{Vector operations}
Take vectors $\vx$, $\vy$
\begin{itemize}
    \item $\vx^\top \vy$: vector inner-product
    \item $\vx \vy^\top$: vector outer-product
\end{itemize}

\paragraph{Operators}
\begin{itemize}
    \item $\odot$: Hadamard product 
    \item $\oplus$: Concatenation
\end{itemize}

\subsection{Implementation Details}\label{appdx:implementation_details}

Most of the experimental details are included across sections~\ref{sec:related_works}-~\ref{sec:results}. Any details which aren't explicitly mentioned either in theses sections or this one are aligned with related works and clearly mentioned as such.

\subsubsection{Weights Initialisation}

\paragraph{$\Phi$} Linear projections, as in $\phi(\cdot)$, were initialised as the identity, following~\cite{zhang2024hedgehog}, with added Gaussian-sampled noise ($\mu=0$ and $\sigma=0.1$) using a seed-based generator to ensure the same initialisation across all runs.

\subsection{Extended Related Works}

\begin{table*}[!hb]
\centering
\small
\renewcommand{\arraystretch}{1.15}
\begin{tabularx}{\textwidth}{l l Y p{0.35\textwidth}}
\toprule
\textbf{Method} & \textbf{Attn Type} & \textbf{Feature map $\phi(x)$} & \textbf{Transfer objective} \\
\midrule
\textbf{T2R/SUPRA} & Fully linear &
$\mathrm{ReLU}(W_\phi^\top x + b)$
&
Uptrain with standard causal LM cross-entropy (no attention-distillation objective).\\

\addlinespace
\textbf{HedgeHog} & Fully linear &
$\exp(W_\phi^\top x + b)$*
&
Minimises cross-entropy / KL between softmax attention weights and linear weights. \\

\addlinespace
\textbf{LoLCATs} & Hybrid (LA\,+\,SWA) &
$\mathrm{Softmax}(W_\phi^\top x + b)$*
&
Minimises MSE between softmax attention outputs and linear (hybrid) attention outputs.\\

\addlinespace
\textbf{Liger} & Hybrid (GLA\,+\,SWA) &
$\mathrm{Softmax}(x)$
&
Causal LM cross-entropy with LoRA fine-tuning (no attention-distillation objective). \\
\bottomrule
\end{tabularx}
\caption{Side-by-side comparison of linearising conversions for pre-trained Transformers. *Methods for which the negative mapping is concatenated to the final output of $\phi$ (see Eq.~\ref{eq:feature_map}).}
\label{tab:linearising-conversions}
\end{table*}

\inappendixtrue
\subsection{Further Results}

\subsubsection{Linear Attention Only}\label{appdx:linear_only}

\begin{table}[!h]
\centering
\resizebox{\linewidth}{!}{
\begin{tabular}{l | c c c c c c | c | c}
\toprule
\textbf{Active Attn Modules} & \textbf{PIQA} & \textbf{ARC-E} & \textbf{ARC-C} & \textbf{HellaSwag} & \textbf{WG} & \textbf{MMLU} & \textbf{AVG} & \textbf{Rec. Perf} \\
\midrule
    \textit{Mistral-7B} & 79.27 & 80.01 & 52.22 & 74.60 & 69.93 & 53.51 & 68.26 & 100.00 \\
    \quad No Attention & 53.92 & 25.63 & 24.66 & 25.99 & 50.67 & \textbf{25.51} & 34.40 & 50.39 \\
    \quad LA Weights Transfer & 67.90 & 61.74 & 29.69 & 45.82 & \textbf{52.64} & 23.82 & 46.94 & 68.76 \\
    \quad \quad +LoRA 1 Epoch & 70.13 & 66.20 & \textbf{34.90} & 56.39 & 51.38 & 24.95 & 50.66 & 74.22 \\
    \quad \quad +LoRA 2 Epochs & \textbf{70.29} & \textbf{66.29} & 34.39 & \textbf{56.41} & 52.17 & 24.97 & \textbf{50.75} & \textbf{74.36} \\

\midrule
    \textit{Llama3-8B} & 78.13 & 81.69 & 56.66 & 75.94 & 71.67 & 63.85 & 71.32 & 100.00 \\
    \quad No Attention & \textbf{55.17} & 26.73 & 22.87 & 26.13 & \textbf{51.14} & 22.95 & 34.17 & 47.90 \\
    \quad LA Weights Transfer & 54.24 & 26.94 & \textbf{26.11} & 26.42 & 50.43 & \textbf{23.42} & 34.59 & 48.50 \\
    \quad \quad +LoRA 1 Epoch & 54.90 & 29.46 & 25.68 & 32.06 & 49.01 & 22.79 & 35.65 & 49.98 \\
    \quad \quad +LoRA 2 Epochs & 54.73 & \textbf{30.26} & 25.43 & \textbf{33.28} & 49.64 & 22.94 & \textbf{36.05} & \textbf{50.54} \\
\midrule
    \textit{Llama3.1-8B} & 80.14 & 81.82 & 55.20 & 79.14 & 73.72 & 68.05 & 73.01 & 100.00 \\
    \quad No Attention & 53.59 & 26.89 & \textbf{25.94} & 26.19 & 49.09 & 22.95 & 34.11 & 46.72 \\
    \quad LA Weights Transfer & 54.41 & 28.16 & 25.34 & 27.50 & 48.46 & \textbf{25.04} & 34.82 & 47.69 \\
    \quad \quad +LoRA 1 Epoch & 58.71 & 38.51 & 24.23 & 35.38 & \textbf{53.51} & 23.53 & 38.98 & 53.39 \\
    \quad \quad +LoRA 2 Epochs & \textbf{58.87} & \textbf{39.69} & 23.89 & \textbf{37.41} & 52.49 & 23.78 & \textbf{39.36} & \textbf{53.90} \\
\bottomrule
\end{tabular}
}
\caption{Measuring the performance of HedgeHog conversion at various points during the conversion process. LoRA refers to a LA-only finetuning.}\label{tab:HedgeHog}
\end{table}

\subsubsection{Attention Transfer Learning Objective}\label{appdx:results_transfer_objective}

\begin{table}[h!]
\centering
\resizebox{\linewidth}{!}{
\begin{tabular}{l | c c c c c c | c | c}
\toprule
\textbf{Transfer Objective} & \textbf{PIQA} & \textbf{ARC-E} & \textbf{ARC-C} & \textbf{HellaSwag} & \textbf{WG} & \textbf{MMLU} & \textbf{AVG} & \textbf{Rec. Perf} \\
\midrule
    \textit{Mistral-7B} & 79.27 & 80.01 & 52.22 & 74.60 & 69.93 & 53.51 & 68.26 & 100.00 \\
    \quad Attention Weights & 67.90 & 61.74 & 29.69 & 45.82 & \textbf{52.64} & 23.82 & 46.94 & 68.76 \\
    \quad Attention Outputs & \textbf{69.64} & \textbf{62.46} & \textbf{31.40} & \textbf{46.61} & 50.28 & 23.00 & \textbf{47.23} & \textbf{69.20} \\
    \quad Hybrid Attention Out. & 55.44 & 32.83 & 24.06 & 27.36 & 49.80 & 22.98 & 35.41 & 51.88 \\
    \quad No Attention & 53.92 & 25.63 & 24.66 & 25.99 & 50.67 & \textbf{25.51} & 34.40 & 50.39 \\
\midrule
    \textit{Llama3-8B} & 78.13 & 81.69 & 56.66 & 75.94 & 71.67 & 63.85 & 71.32 & 100.00 \\
    \quad Attention Weights & 54.24 & 26.94 & \textbf{26.11} & 26.42 & 50.43 & 23.42 & 34.59 & 48.50 \\
    \quad Attention Outputs & \textbf{61.04} & \textbf{41.04} & 24.06 & \textbf{30.02} & 49.25 & \textbf{23.88} & \textbf{38.22} & \textbf{53.58} \\
    \quad Hybrid Attention Out. & 53.10 & 26.22 & 24.74 & 26.23 & 49.41 & 23.43 & 33.86 & 47.47 \\
    \quad No Attention & 55.17 & 26.73 & 22.87 & 26.13 & \textbf{51.14} & 22.95 & 34.17 & 47.90 \\
\midrule
    \textit{Llama3.1-8B} & 80.14 & 81.82 & 55.20 & 79.14 & 73.72 & 68.05 & 73.01 & 100.00 \\
    \quad Attention Weights & 54.41 & 28.16 & 25.34 & 27.50 & 48.46 & 25.04 & 34.82 & 47.69 \\
    \quad Attention Outputs & \textbf{54.90} & \textbf{28.49} & 25.60 & \textbf{28.60} & 48.78 & \textbf{26.30} & \textbf{35.45} & \textbf{48.55} \\
    \quad Hybrid Attention Out. & 54.57 & 25.46 & \textbf{26.02} & 26.04 & 47.75 & 23.37 & 33.87 & 46.39 \\
    \quad No Attention & 53.59 & 26.89 & 25.94 & 26.19 & \textbf{49.09} & 22.95 & 34.11 & 46.72 \\
\bottomrule
\end{tabular}
}
\caption{Comparing the performance of linear attention only after the attention transfer phase across learning objectives used during this stage.}\label{tab:TransferObjective_all}
\end{table}

\subsubsection{Adding SWA at Inference Time}\label{appdx:results_inference_time_hybrid}

\begin{table}[h!]
\centering
\resizebox{\linewidth}{!}{
\begin{tabular}{l | c c c c c c | c | c}
\toprule
\textbf{Active Attn Modules} & \textbf{PIQA} & \textbf{ARC-E} & \textbf{ARC-C} & \textbf{HellaSwag} & \textbf{WG} & \textbf{MMLU} & \textbf{AVG} & \textbf{Rec. Perf} \\
\midrule
    \textit{Mistral-7B} & 79.27 & 80.01 & 52.22 & 74.60 & 69.93 & 53.51 & 68.26 & 100.00 \\
    \quad LA Weights Transfer & 67.90 & 61.74 & 29.69 & 45.82 & 52.64 & 23.82 & 46.94 & 68.76 \\
    \quad \quad \textit{+SWA}& \textbf{79.11} & 79.76 & 51.62 & \textbf{70.57} & 69.93 & \textbf{45.13} & \textbf{66.02} & \textbf{96.72} \\
    \quad \quad \textit{+SWA, Overlap} & 77.48 & 78.79 & 50.77 & 67.13 & 61.96 & 43.88 & 63.34 & 92.79 \\
    \quad \quad \textit{SWA-only} & 77.97 & \textbf{79.92} & 51.02 & 51.19 & 69.93 & 39.92 & 61.66 & 90.33 \\
    \addlinespace
    \quad +LoRA 1 Epoch & 70.13 & 66.20 & 34.90 & 56.39 & 51.38 & 24.95 & 50.66 & 74.22 \\
    \quad \quad \textit{+SWA} & 78.73 & 78.83 & \textbf{51.96} & 70.44 & \textbf{70.24} & 42.66 & 65.48 & 95.93 \\
    \quad \quad \textit{+SWA, Overlap} & 77.42 & 78.20 & 49.06 & 67.38 & 63.30 & 40.84 & 62.70 & 91.86 \\
    \quad \quad \textit{SWA-only} & 77.91 & 78.75 & 51.54 & 48.63 & \textbf{70.24} & 39.41 & 61.08 & 89.49 \\
    \addlinespace
    \quad +LoRA 2 Epochs & 70.29 & 66.29 & 34.39 & 56.41 & 52.17 & 24.97 & 50.75 & 74.36 \\
    \quad \quad \textit{+SWA} & 78.62 & 78.70 & 51.11 & 69.75 & 69.61 & 42.14 & 64.99 & 95.21 \\
    \quad \quad \textit{+SWA, Overlap} & 77.26 & 77.78 & 48.04 & 66.22 & 62.98 & 39.99 & 62.05 & 90.90 \\
    \quad \quad \textit{SWA-only} & 77.80 & 78.49 & 50.77 & 47.69 & 69.61 & 38.91 & 60.55 & 88.70 \\
\midrule
    \textit{Llama3-8B} & 78.13 & 81.69 & 56.66 & 75.94 & 71.67 & 63.85 & 71.32 & 100.00 \\
    \quad LA Weights Transfer & 54.24 & 26.94 & 26.11 & 26.42 & 50.43 & 23.42 & 34.59 & 48.50 \\
    \quad \quad \textit{+SWA} & 77.80 & \textbf{81.06} & \textbf{55.97} & 71.71 & 71.67 & 49.22 & 67.91 & 95.21 \\
    \quad \quad \textit{+SWA, Overlap} & 67.03 & 56.06 & 36.86 & 49.67 & 58.56 & 40.26 & 51.41 & 72.08 \\
    \quad \quad \textit{SWA-only} & 76.28 & 80.09 & 55.12 & 44.48 & 71.67 & 43.31 & 61.83 & 86.68 \\
    \addlinespace
    \quad +LoRA 1 Epoch & 54.90 & 29.46 & 25.68 & 32.06 & 49.01 & 22.79 & 35.65 & 49.98 \\
    \quad \quad \textit{+SWA} & 77.80 & \textbf{81.06} & \textbf{55.97} & 71.71 & 71.67 & 49.22 & 67.91 & 95.90 \\
    \quad \quad \textit{+SWA, Overlap} & 67.03 & 56.06 & 36.86 & 49.67 & 58.56 & 40.26 & 51.41 & 78.42 \\
    \quad \quad \textit{SWA-only} & 76.28 & 80.09 & 55.12 & 44.48 & 71.67 & 43.31 & 61.83 & 87.20 \\
    \addlinespace 
    \quad +LoRA 2 Epochs & 54.73 & 30.26 & 25.43 & 33.28 & 49.64 & 22.94 & 36.05 & 50.54 \\
    \quad \quad \textit{+SWA} & \textbf{78.35} & 80.22 & 55.55 & \textbf{73.15} & \textbf{73.24} & \textbf{49.87} & \textbf{68.40} & \textbf{95.27} \\
    \quad \quad \textit{+SWA, Overlap} & 70.62 & 61.20 & 41.13 & 58.61 & 60.30 & 43.71 & 55.93 & 78.10 \\
    \quad \quad \textit{SWA-only} & 76.99 & 79.46 & 54.01 & 45.71 & \textbf{73.24} & 43.77 & 62.20 & 86.74 \\
\midrule
    \textit{Llama3.1-8B} & 80.14 & 81.82 & 55.20 & 79.14 & 73.72 & 68.05 & 73.01 & 100.00 \\
    \quad LA Weights Transfer & 54.41 & 28.16 & 25.34 & 27.50 & 48.46 & 25.04 & 34.82 & 47.69 \\
    \quad \quad \textit{+SWA} & 79.87 & 80.98 & \textbf{54.78} & \textbf{72.81} & 73.72 & 50.11 & 68.71 & 94.11 \\
    \quad \quad \textit{+SWA, Overlap} & 71.38 & 64.35 & 40.44 & 51.66 & 62.51 & 45.90 & 56.04 & 76.75 \\
    \quad \quad \textit{SWA-only} & 78.51 & 80.35 & 53.24 & 46.08 & 73.72 & 43.51 & 62.57 & 85.70 \\
    \addlinespace
    \quad +LoRA 1 Epoch & 58.71 & 38.51 & 24.23 & 35.38 & 53.51 & 23.53 & 38.98 & 53.39 \\
    \quad \quad \textit{+SWA} & \textbf{80.09} & \textbf{81.44} & 53.58 & 72.77 & \textbf{74.03} & \textbf{50.76} & \textbf{68.78} & \textbf{94.20} \\
    \quad \quad \textit{+SWA, Overlap} & 76.88 & 75.93 & 48.89 & 61.16 & 62.59 & 47.79 & 62.21 & 85.20 \\
    \quad \quad \textit{SWA-only} & 78.94 & 80.64 & 51.96 & 46.24 & \textbf{74.03} & 43.64 & 62.58 & 85.71 \\
    \addlinespace
    \quad +LoRA 2 Epochs & 58.87 & 39.69 & 23.89 & 37.41 & 52.49 & 23.78 & 39.36 & 53.90 \\
    \quad \quad \textit{+SWA} & 79.65 & 80.72 & 53.58 & 72.07 & 73.40 & 49.56 & 68.16 & 93.36 \\
    \quad \quad \textit{+SWA, Overlap} & 77.48 & 77.82 & 50.51 & 61.46 & 63.38 & 47.98 & 63.11 & 86.43 \\
    \quad \quad \textit{SWA-only} & 78.45 & 79.92 & 52.39 & 45.27 & 73.40 & 43.58 & 62.17 & 85.15 \\
\bottomrule
\end{tabular}
}
\caption{Adding SWA at inference time at various points of the conversion process.}\label{tab:zero_shot_hybrid}
\end{table}

\subsubsection{HedgeCATs}\label{appdx:results_hedgecats}
\begin{table}[h!]
\centering
\resizebox{\linewidth}{!}{
\begin{tabular}{l | c c c c c c c c}
\toprule
\textbf{Active Attn Module} & \textbf{PIQA} & \textbf{ARC-E} & \textbf{ARC-C} & \textbf{HellaSwag} & \textbf{WG} & \textbf{MMLU} & \textbf{AVG} & \textbf{Rec. Perf} \\
\midrule
    \textit{Mistral-7B} & 79.27 & 80.01 & 52.22 & 74.60 & 69.93 & 53.51 & 68.26 & 100.00 \\
    \quad 1 epoch  &  &  &  &  &  &  &  & \\
    \quad \quad Linear+SWA & 78.51 & 79.46 & \textbf{51.45} & \textbf{71.14} & \textbf{69.38} & 45.14 & \textbf{65.85} & \textbf{96.47} \\ 
    \quad \quad SWA-only & 78.24 & 79.34 & 50.68 & 65.92 & \textbf{69.38} & 46.96 & 65.09 & 95.36 \\ 
    \quad 2 epochs &  &  &  &  &  &  &  & \\
    \quad \quad Linear+SWA & 78.24 & 79.59 & 51.02 & 70.73 & 68.75 & 45.24 & 65.60 & 96.10 \\ 
    \quad \quad SWA-only & 78.07 & 79.71 & 50.17 & 66.30 & 68.75 & 47.07 & 65.01 & 95.25 \\ 
    \quad 5 epochs &  &  &  &  &  &  &  &  \\
    \quad \quad Linear+SWA & \textbf{78.56} & 79.42 & 50.26 & 70.55 & 67.64 & 45.62 & 65.34 & 95.73 \\ 
    \quad \quad SWA-only & 78.51 & \textbf{79.67} & 49.66 & 66.51 & 67.64 & \textbf{47.41} & 64.90 & 95.08 \\ 

\midrule
    \textit{Llama3-8B} & 78.13 & 81.69 & 56.66 & 75.94 & 71.67 & 63.85 & 71.32 & 100.00 \\
    \quad 1 epoch &  &  &  &  &  &  &  & \\
    \quad \quad Linear+SWA & \textbf{79.38} & \textbf{81.78} & \textbf{55.97} & \textbf{72.28} & 73.32 & 51.18 & \textbf{68.99} & \textbf{96.72} \\ 
    \quad \quad \quad SWA-only & 78.89 & 81.61 & 55.63 & 59.67 & 73.32 & 47.02 & 66.02 & 92.57 \\ 
    \quad 2 epochs &  &  &  &  &  &  &  & \\
    \quad \quad Linear+SWA & 78.62 & 80.89 & 53.92 & 71.54 & \textbf{73.48} & 51.87 & 68.39 & 95.88 \\ 
    \quad \quad \quad SWA-only & 78.40 & 81.14 & 54.27 & 62.68 & \textbf{73.48} & 47.76 & 66.29 & 92.94 \\
    \quad 5 epochs  &  &  &  &  &  &  &  & \\ 
    \quad \quad Linear+SWA & 78.73 & 80.05 & 52.99 & 71.49 & 72.38 & \textbf{52.90} & 68.09 & 95.47 \\
    \quad \quad \quad SWA-only & 78.78 & 80.39 & 53.33 & 65.57 & 72.38 & 47.76 & 66.37 & 93.05 \\ 
\end{tabular}
}
\caption{HedgeCATs performance for different activate attention modules at inference time and at different stages of LoRA finetuning.}
\label{tab:hedgecats}
\end{table}

\subsubsection{SSD: Testing Scheduled Hybrid Conversion}\label{appdx:results_ssd}
\begin{table}[h!]
\centering
\resizebox{\linewidth}{!}{
\begin{tabular}{l | c c c c c c c c}
\toprule
\textbf{Active Attn Modules} & \textbf{PIQA} & \textbf{ARC-E} & \textbf{ARC-C} & \textbf{HellaSwag} & \textbf{WG} & \textbf{MMLU} & \textbf{AVG} & \textbf{Rec. Perf} \\
\midrule
    \textit{Mistral-7B} & 79.27 & 80.01 & 52.22 & 74.60 & 69.93 & 53.51 & 68.26 & 100.00 \\
    \quad 1 epoch  &  &  &  &  &  &  &  & \\
    \quad \quad Linear+SWA & \textbf{78.94} & 79.55 & 52.47 & 71.55 & 69.61 & 44.79 & 66.15 & 96.92 \\ 
    \quad \quad + Overlap & 77.86 & 78.58 & 50.34 & 68.46 & 63.61 & 42.79 & 63.61 & 93.19  \\ 
    \quad \quad SWA-only  & 78.24 & 79.59 & 52.39 & 53.77 & 69.61 & 40.74 & 62.39 & 91.40 \\ 
    \quad 2 epochs &  &  &  &  &  &  &  & \\
    \quad \quad Linear+SWA & 78.62 & \textbf{79.59} & 52.90 & \textbf{71.61} & 68.98 & \textbf{45.13} & 66.14 & 96.90 \\ 
    \quad \quad + Overlap & 77.64 & 78.49 & 50.43 & 68.55 & 64.25 & 42.80 & 63.69 & 93.31 \\ 
    \quad \quad SWA-only  & 77.80 & 79.76 & 52.39 & 55.45 & 68.98 & 41.02 & 62.57 & 91.66 \\ 
    \quad 5 epochs &  &  &  &  &  &  &  & \\
    \quad \quad Linear+SWA &  78.62 & 79.55 & \textbf{53.58} & 71.06 & \textbf{69.93} & 45.04 & \textbf{66.30} & \textbf{97.13} \\ 
    \quad \quad + Overlap & 77.48 & 77.90 & 50.68 & 68.09 & 62.35 & 42.24 & 63.12 & 92.48 \\ 
    \quad \quad SWA-only  & 78.40 & 79.25 & 53.24 & 61.45 & 69.93 & 41.97 & 64.04 & 93.82 \\ 

\midrule
    \textit{Llama3-8B} & 78.13 & 81.69 & 56.66 & 75.94 & 71.67 & 63.85 & 71.32 & 100.00 \\
    \quad 1 epoch &  &  &  &  &  &  &  & \\
    \quad \quad Linear+SWA & \textbf{78.40} & 80.72 & 55.72 & 72.07 & 71.43 & 50.38 & 68.12 & 95.59 \\ 
    \quad \quad \quad + Overlap & 70.89 & 63.80 & 41.81 & 63.53 & 59.98 & 47.98 & 58.00 & 81.32 \\ 
    \quad \quad \quad SWA-only  & 77.26 & 79.88 & 54.61 & 48.04 & 71.43 & 44.70 & 62.65 & 87.84 \\ 
    \quad 2 epochs &  &  &  &  &  &  &  & \\
    \quad \quad Linear+SWA  & 78.18 & \textbf{80.85} & 55.80 & 72.34 & 72.22 & 50.34 & 68.29 & 95.74 \\ 
    \quad \quad \quad + Overlap & 71.33 & 63.64 & 41.47 & 65.52 & 59.98 & 48.15 & 58.35 & 82.29 \\ 
    \quad \quad \quad SWA-only  & 77.15 & 80.05 & 54.61 & 50.31 & 72.22 & 45.28 & 63.27 & 88.38 \\ 
    \quad 5 epochs  &  &  &  &  &  &  &  & \\ 
    \quad \quad Linear+SWA &  78.13 & 80.72 & \textbf{56.23} & \textbf{72.84} & \textbf{72.61} & \textbf{50.97} & \textbf{68.58} & \textbf{95.97} \\
    \quad \quad \quad + Overlap & 71.55 & 63.30 & 41.13 & 68.13 & 60.38 & 49.75 & 59.04 & 82.78 \\ 
    \quad \quad \quad SWA-only  & 77.48 & 80.43 & 55.12 & 56.11 & 72.61 & 46.16 & 64.65 & 90.65 \\ 
\end{tabular}
}
\caption{Finetuning using SSD for a dropout schedule of 0.9, 0.75, 0.5 and a fixed sliding window size of 32.}\label{tab:dropout_schedule_SW32}
\end{table}

\begin{table}[h!]
\centering
\resizebox{\linewidth}{!}{
\begin{tabular}{l | c c c c c c c c}
\toprule
\textbf{Active Attn Modules} & \textbf{PIQA} & \textbf{ARC-E} & \textbf{ARC-C} & \textbf{HellaSwag} & \textbf{WG} & \textbf{MMLU} & \textbf{AVG} & \textbf{Rec. Perf} \\
\midrule
    \textit{Mistral-7B} & 79.27 & 80.01 & 52.22 & 74.60 & 69.93 & 53.51 & 68.26 & 100.00 \\
    \quad 1 epoch &  &  &  &  &  &  &  & \\
    \quad \quad Linear+SWA & 78.24 & 77.95 & 50.34 & 67.30 & 68.27 & 43.93 & 64.34 & 94.26 \\ 
        \quad \quad + Overlap  & 77.37 & 77.19 & 47.10 & 66.63 & 59.91 & 42.36 & 61.76 & 90.48\\ 
        \quad \quad SWA-only  & 77.69 & 78.28 & 50.00 & 58.59 & 68.27 & 41.65 & 62.41 & 91.44 \\ 
    \quad 2 epochs &  &  &  &  &  &  &  & \\
    \quad \quad Linear+SWA & \textbf{78.78} & 79.34 & \textbf{53.33} & \textbf{70.47} & \textbf{69.06} & \textbf{44.95} & \textbf{65.99} & \textbf{96.68} \\ 
        \quad \quad + Overlap & 77.31 & 77.44 & 48.46 & 66.74 & 61.48 & 43.04 & 62.41 & 91.44 \\ 
        \quad \quad SWA-only  & 78.29 & \textbf{79.46} & 52.56 & 61.13 & 69.06 & 42.40 & 63.82 & 93.50 \\ 
    \quad 5 epochs &  &  &  &  &  &  &  & \\
    \quad \quad Linear+SWA & 78.51 & 79.12 & 52.13 & 68.30 & \textbf{69.06} & 42.69 & 64.97 & 95.18 \\ 
        \quad \quad + Overlap & 76.82 & 76.52 & 47.95 & 62.36 & 60.54 & 39.30 & 60.58 & 88.76 \\ 
        \quad \quad SWA-only  & 78.18 & 79.08 & 51.62 & 62.49 & \textbf{69.06} & 41.52 & 63.66 & 93.26 \\ 
\midrule
    \textit{Llama3-8B} & 78.13 & 81.69 & 56.66 & 75.94 & 71.67 & 63.85 & 71.32 & 100.00 \\
    \quad 1 epoch &  &  &  &  &  &  &  & \\
    \quad \quad Linear+SWA & 78.07 & 80.64 & 55.38 & 72.23 & \textbf{72.38} & 50.38 & 68.18 & 95.59 \\ 
        \quad \quad + Overlap & 71.16 & 65.03 & 42.15 & 63.66 & 60.22 & 48.08 & 58.38 & 81.86 \\ 
        \quad \quad SWA-only  &  76.88 & 79.76 & 54.27 & 48.21 & 72.38 & 44.59 & 62.68 & 87.88 \\ 
    \quad 2 epochs &  &  &  &  &  &  &  & \\
    \quad \quad Linear+SWA &  78.07 & \textbf{80.77} & 54.86 & 72.41 & 71.59 & 50.46 & 68.03 & 95.38 \\ 
        \quad \quad + Overlap &71.38 & 64.56 & 41.55 & 65.71 & 60.77 & 48.20 & 58.70 & 82.29  \\ 
        \quad \quad SWA-only  &77.04 & 80.05 & 53.75 & 50.37 & 71.59 & 45.41 & 63.04 & 88.38 \\ 
    \quad 5 epochs &  &  &  &  &  &  &  & \\
        \quad \quad Linear+SWA & \textbf{78.35} & 80.43 & \textbf{55.97} & \textbf{72.89} & 72.14 & \textbf{50.90} & \textbf{68.45} & \textbf{95.97} \\ 
        \quad \quad + Overlap & 71.38 & 63.47 & 42.06 & 68.12 & 60.30 & 49.65 & 59.16 & 82.95 \\ 
        \quad \quad SWA-only & 77.64 & 80.18 & 54.86 & 56.07 & 72.14 & 46.20 & 64.52 & 90.45 \\ 
\end{tabular}
}
\caption{Finetuning using SSD for a fixed dropout of 0.5 and sliding window size of 16.}\label{tab:swa_schedule_all_fixed}
\end{table}

\begin{table}[h!]
\centering
\resizebox{\linewidth}{!}{
\begin{tabular}{l | c c c c c c c c}
\toprule
\textbf{Active Attn Modules} & \textbf{PIQA} & \textbf{ARC-E} & \textbf{ARC-C} & \textbf{HellaSwag} & \textbf{WG} & \textbf{MMLU} & \textbf{AVG} & \textbf{Rec. Perf} \\
\midrule
    \textit{Mistral-7B} & 79.27 & 80.01 & 52.22 & 74.60 & 69.93 & 53.51 & 68.26 & 100.00 \\
    \quad 1 epoch &  &  &  &  &  &  &  & \\
    \quad \quad Linear+SWA & \textbf{77.75} & 77.74 & \textbf{50.17} & \textbf{66.81} & \textbf{68.43} & \textbf{43.28} & \textbf{64.03} & \textbf{93.81} \\ 
        \quad \quad + Overlap & 77.26 & 76.56 & 46.67 & 66.01 & 59.91 & 42.24 & 61.44 & 90.02 \\ 
        \quad \quad SWA-only  & 77.42 & \textbf{77.99} & 49.74 & 58.31 & \textbf{68.43} & 41.71 & 62.27 & 91.22 \\ 
    \quad 2 epochs &  &  &  &  &  &  &  & \\
    \quad \quad Linear+SWA & 76.82 & 74.07 & 46.08 & 66.44 & 68.43 & 42.73 & 62.43 & 91.46  \\ 
        \quad \quad + Overlap & 75.84 & 72.31 & 40.96 & 65.79 & 58.88 & 41.48 & 59.21 & 86.75   \\ 
        \quad \quad SWA-only  & 76.82 & 74.71 & 46.50 & 58.98 & 68.43 & 41.70 & 61.19 & 89.65 \\ 
    \quad 5 epochs & & & & & & & &  \\
    \quad \quad Linear+SWA & 76.33 & 73.65 & 46.08 & 63.96 & 67.17 & 41.34 & 61.42 & 89.99 \\ 
        \quad \quad + Overlap & 75.63 & 72.31 & 42.41 & 62.89 & 59.27 & 38.66 & 58.53 & 85.75  \\ 
        \quad \quad SWA-only  & 76.22 & 73.91 & 46.25 & 58.38 & 67.17 & 41.01 & 60.49 & 88.62 \\ 
\midrule
    \textit{Llama3-8B} & 78.13 & 81.69 & 56.66 & 75.94 & 71.67 & 63.85 & 71.32 & 100.00 \\
    \quad 1 epoch &  &  &  &  &  &  &  & \\
    \quad \quad Linear+SWA  & 77.91 & \textbf{80.39} & \textbf{54.27}  & 71.49 & 72.06 & \textbf{51.00} & \textbf{67.85} & \textbf{95.13}  \\
    \quad \quad + Overlap &  71.00 & 63.64 & 40.87 & 66.37 & 61.40 & 48.88 & 58.69 & 82.29  \\ 
    \quad \quad SWA-only  & 77.26 & 80.13 & 53.75 & 55.03 & 72.06 & 45.93 & 64.03 & 89.77 \\ 
    \quad 2 epochs &  &  &  &  &  &  &  & \\
    \quad \quad Linear+SWA & 77.48 & 79.50 & 53.67 & 71.63 & \textbf{72.45} & 50.35 & 67.51 & 94.66  \\
    \quad \quad + Overlap & 70.08 & 60.23 & 39.08 & 66.63 & 60.93 & 48.06 & 57.50 & 80.62  \\ 
    \quad \quad SWA-only  & 77.31 & 79.34 & 53.16 & 56.75 & 72.45 & 45.64 & 64.11 & 89.88  \\ 
    \quad 5 epochs &  &  &  &  &  &  &  & \\
    \quad \quad Linear+SWA & \textbf{78.02} & 79.59 & 54.10 & \textbf{71.65} & 71.67 & 49.67 & 67.45 & 94.57 \\
    \quad \quad + Overlap & 68.82 & 58.50 & 38.14 & 66.24 & 58.88 & 46.99 & 56.26 & 78.88 \\ 
    \quad \quad SWA-only  & 77.37 & 79.59 & 52.99 & 57.16 & 71.67 & 45.37 & 64.03 & 89.77 \\ 
\bottomrule
\end{tabular}
}
\caption{Finetuning using SSD for a fixed dropout of 0.5 and a sliding window size schedule of 4, 8, 16, 32, 64.}\label{tab:swa_schedule_fixed_dropout}
\end{table}
\label{sec:appendix}
\FloatBarrier\clearpage
\inappendixfalse

\end{document}